\def\year{2023}\relax
\documentclass[letterpaper]{article} 
\usepackage{aaai24}  
\usepackage{epsfig}
\usepackage{amsmath}
\usepackage{amssymb}
\usepackage{cite}
\usepackage{amsthm}
\usepackage{mathrsfs}
\usepackage{booktabs}  
\usepackage{times}  
\usepackage{helvet}  
\usepackage{courier}  
\usepackage[hyphens]{url}  
\usepackage{graphicx} 
\usepackage{natbib}  
\usepackage{caption} 
\DeclareCaptionStyle{ruled}{labelfont=normalfont,labelsep=colon,strut=off} 
\usepackage{algorithm}
\usepackage{algorithmic}

\usepackage{multirow}

\usepackage{newfloat}
\usepackage{listings}
\floatstyle{ruled}
\newfloat{listing}{tb}{lst}{}
\floatname{listing}{Listing}
\title{\emph{AvatarVerse}: High-quality \& Stable 3D Avatar Creation from Text and Pose}

\author{
    Huichao Zhang\textsuperscript{\rm 1}\equalcontrib, Bowen Chen\textsuperscript{\rm 1}\equalcontrib, Hao Yang\textsuperscript{\rm 1}, Liao Qu\textsuperscript{\rm 1, 2}, Xu Wang\textsuperscript{\rm 1}\\
    Li Chen\textsuperscript{\rm 1}, Chao Long\textsuperscript{\rm 1}, Feida Zhu\textsuperscript{\rm 1}, Kang Du\textsuperscript{\rm 1}, Min Zheng\textsuperscript{\rm 1}\\
}
\affiliations{
    \textsuperscript{\rm 1}ByteDance, Beijing, China. \\
    \textsuperscript{\rm 2}Department of Electrical and Computer Engineering, Carnegie Mellon University, PA, USA.\\
    \{zhanghuichao.hc, chenbowen.cbw, wangxu.ailab, chenli.phd, longchao, zhufeida, dukang.daniel, zhengmin.666\}@bytedance.com, liaoq@andrew.cmu.edu, yanghao.alexis@foxmail.com
}

\begin{document}
\maketitle

\begin{figure*}[htbp]
\begin{center}
\includegraphics[width=0.95\linewidth]{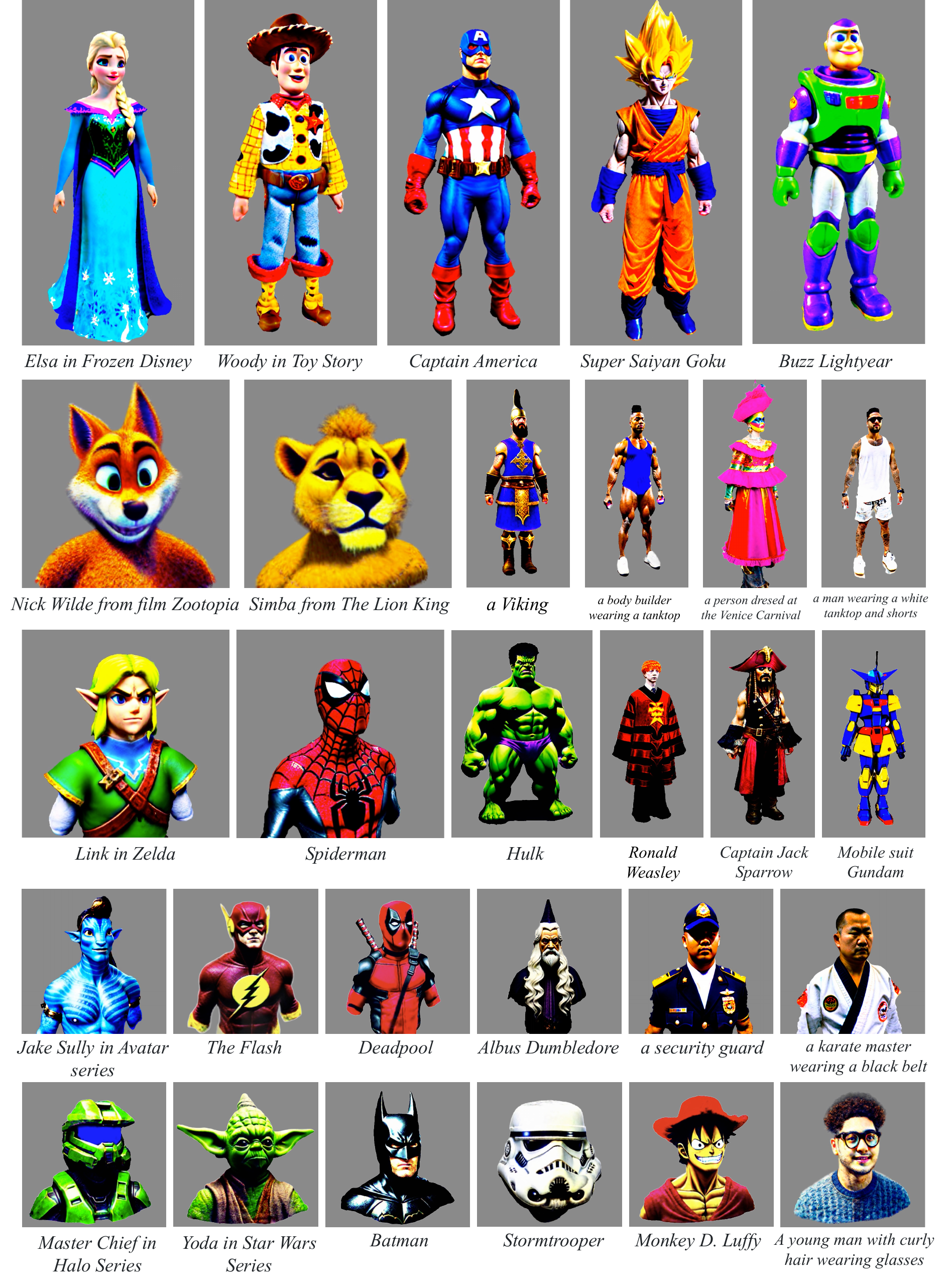}
\end{center}
\caption{High-quality 3D avatars generated by AvatarVerse based on a simple text description.}
\label{fig:big}
\end{figure*}

\begin{abstract}

Creating expressive, diverse and high-quality 3D avatars from highly customized text descriptions and pose guidance is a challenging task, due to the intricacy of modeling and texturing in 3D that ensure details and various styles (realistic, fictional, etc). 
We present \emph{AvatarVerse}, a stable pipeline for generating expressive high-quality 3D avatars from nothing but text descriptions and pose guidance. 
In specific, we introduce a 2D diffusion model conditioned on DensePose signal to establish 3D pose control of avatars through 2D images, which enhances view consistency from partially observed scenarios. It addresses the infamous Janus Problem and significantly stablizes the generation process.
Moreover, we propose a progressive high-resolution 3D synthesis strategy, which obtains substantial improvement over the quality of the created 3D avatars.
To this end, the proposed AvatarVerse pipeline achieves zero-shot 3D modeling of 3D avatars that are not only more expressive, but also in higher quality and fidelity than previous works.
Rigorous qualitative evaluations and user studies showcase AvatarVerse's superiority in synthesizing high-fidelity 3D avatars, leading to a new standard in high-quality and stable 3D avatar creation.
Our project page is: \url{https://avatarverse3d.github.io/} .

\end{abstract}

\section{1. Introduction}\label{sec:1}

The creation of high-quality 3D avatars has garnered significant interest due to their widespread applications in domains such as game production, social media and communication, augmented and virtual reality (AR/VR), and human-computer interaction. 
Traditional manual construction of these intricate 3D models is a labor-intensive and time-consuming process, requiring thousands of hours from skilled artists possessing extensive aesthetic and 3D modeling expertise.
Consequently, automating the generation of high-quality 3D avatars using only natural language descriptions holds great research prospects with the potential to save resources, which is also the goal of our work.

In recent years, significant efforts have been made in reconstructing high-fidelity 3D avatars from multi-view videos \cite{Jiang2022InstantAvatarLA, Li2023PoseVocabLJ, Zheng2023AvatarReXRE, Wang2023StyleAvatarRP, Isik2023HumanRFHN} or reference images \cite{Wang2021CrossDomainAD, Xiu2022ECONEC}. 
These methods primarily rely on limited visual priors sourced from videos or reference images, leading to constrained ability to generate creative avatars with complex text prompts. 
In 2D image generation, diffusion models \cite{Rombach2021HighResolutionIS, Zhang2023AddingCC, Saharia2022PhotorealisticTD} illustrate considerable creativity, primarily due to the availability of large-scale text-image pairs. Nevertheless, the scarcity and limited diversity of 3D models present challenges to effectively training a 3D diffusion model. 
Recent studies \cite{Poole2022DreamFusionTU, Cao2023DreamAvatarTG, Huang2023DreamWaltzMA, Kolotouros2023DreamHumanA3} have investigated the use of pre-trained text-image generative models to optimize Neural Radiance Fields (NeRF) \cite{Mildenhall2020NeRFRS} for generating high-fidelity 3D models. Yet, stable creation of 
high-quality 3D avatars exhibiting various poses, appearances, and shapes remains a difficult task. 
For example, employing common score distillation sampling (SDS) \cite{Poole2022DreamFusionTU} to guide NeRF optimization without additional control tends to bring in the Janus (multi-face) problem.
Also, the avatars produced by current approaches tend to exhibit noticeable blurriness and coarseness, leading to the absence of high-resolution local texture details, accessories, and other relevant features.

To cope with these weaknesses, we propose \emph{AvatarVerse}, a novel framework designed for generating high-quality and stable 3D avatars from textual descriptions and pose guidances. 
We first train a new ControlNet with human DensePose condition \cite{Gler2018DensePoseDH} over 800K images. SDS loss conditinal on the 2D DensePose signal is then implemented on top of the ControlNet. Through this way, we obtain precise view correspondence between different 2D views as well as between every 2D view and the 3D space. Our approach not only enables pose control of the generated avatars, but also eliminates the Janus Problem suffered by most existing methods. It thus ensures a more stable and view-consistent avatar creation process.
Additionally, benefiting from the accurate and flexible supervision signals provided by DensePose, the generated avatars can be highly aligned with the joints of the SMPL model, enabling simple and effective skeletal binding and control. 

While relying solely on DensePose-conditioned ControlNet may result in local artifacts, we introduce a progressive high-resolution generation strategy to enhance the fidelity and detail of local geometry. To alleviate the coarseness of the generated avatar, we incorporate a smoothness loss, which regularizes the synthesis procedure by encouraging a smoother gradient of the density voxel grid within our computationally efficient explicit Neural Radiance Fields (NeRF).

The overall contributions are as follows:

\begin{itemize}
\item[$\bullet$] We present \emph{AvatarVerse}, a method that can automatically create a high-quality 3D avatar accoding to nothing but a text description and a reference human pose.

\item[$\bullet$] We present the DensePose-Conditioned Score Distillation Sampling Loss, an approach that facilitates pose-aware 3D avatar synthesis and effectively mitigates the Janus problem, thereby enhancing system stability.

\item[$\bullet$] We bolster the quality of the produced 3D avatars via a progressive high-resolution generation strategy. This method, through a meticulous coarse-to-fine refining process, synthesizes 3D avatars with superior detail, encompassing elements like hands, accessories, and beyond.

\item[$\bullet$] AvatarVerse delivers exceptional performance, excelling in both quality and stability. Rigorous qualitative evaluations, complemented by comprehensive user studies, underscore AvatarVerse's supremacy in crafting high-fidelity 3D avatars, thereby setting a new benchmark in stable, zero-shot 3D avatar creation of the highest quality.

\end{itemize}

\section{2. Related work} \label{sec: 2}

\subsection{2.1. Text-guided 3D content generation}

The success in text-guided 2D image generation has paved the way for the development of text-guided 3D content generation methods. CLIP-forge \cite{Sanghi2021CLIPForgeTZ}, DreamFields \cite{Jain2021ZeroShotTO}, and CLIP-Mesh \cite{Khalid2022CLIPMeshGT} utilize the CLIP model \cite{Radford2021LearningTV} to optimize underlying 3D representations such as meshes and NeRF. DreamFusion \cite{Poole2022DreamFusionTU} first proposes score distillation sampling (SDS) loss to get supervision from a pre-trained diffusion model \cite{Saharia2022PhotorealisticTD} during the 3D generation. Latent-NeRF \cite{metzer2022latent} improves upon DreamFusion by optimizing a NeRF that operates the diffusion process in a latent space. TEXTure \cite{Richardson2023TEXTureTT} generates texture maps using a depth diffusion model for a given 3D mesh. ProlificDreamer \cite{Wang2023ProlificDreamerHA} proposes variational score distillation and produces high-resolution and high-fidelity results. Despite their promising performance in 3D general content generation, these methods often produce suboptimal results when generating avatars, exhibiting issues like low quality, Janus (multi-face) problem, and incorrect body parts. In contrast, our AvatarVerse enables an accurate and high-quality generation of 3D avatars from text prompts.

\subsection{2.2. Text-guided 3D Avatar generation}
Avatar-CLIP \cite{Hong2022AvatarCLIPZT} first initializes 3D human geometry with a shape VAE network and utilizes CLIP \cite{Radford2021LearningTV} to facilitate geometry sculpting and texture generation.
DreamAvatar \cite{Cao2023DreamAvatarTG} and AvatarCraft \cite{Jiang2023AvatarCraftTT} employ the SMPL model as a shape prior and utilize pretrained text-to-image diffusion models to generate 3D avatars. DreamFace \cite{Zhang2023DreamFacePG} introduces a coarse-to-fine scheme to create personalized 3D facial structures.
HeadSculpt \cite{Han2023HeadSculptC3} generates 3D head avatars by leveraging landmark-based control and a learned textual embedding representing the back view appearance of heads.
Concurrent with our work, DreamWaltz \cite{Huang2023DreamWaltzMA} presents 3D-consistent occlusion-aware score distillation sampling, which incorporates 3D-aware skeleton conditioning for view-aligned supervision. Constrained by the original training data, the skeleton-conditioned diffusion model may still exhibit view inconsistencies such as failing to generate the backside of desired avatars or struggling to generate specific body parts when provided with partial skeleton information. Furthermore, the sparse nature of the skeleton makes it challenging for the model to determine avatar contours and edges, leading to low-quality results. On the contrary, our proposed DensePose-conditioned ControlNet ensures high-quality, view-consistent image generation of various viewpoints and body parts, including full body, legs, head, and more, guaranteeing superior avatar quality. 

\subsection{2.3. High-quality 3D Avatar Generation}
Recently, there has been a growing focus on achieving high-quality or high-fidelity 3D generation and reconstruction.
Some methods attempt to generate high-fidelity 3D human avatars from multi-view RGB videos \cite{Jiang2022InstantAvatarLA, Li2023PoseVocabLJ, Zheng2023AvatarReXRE, Wang2023StyleAvatarRP, Isik2023HumanRFHN}. There has also been work \cite{Lin2022Magic3DHT} explored a coarse-to-fine methodology, specifically by optimizing a high-resolution latent diffusion model to refine a textured 3D mesh model. 
In parallel to our work, DreamHuman \cite{Kolotouros2023DreamHumanA3} zooms in and renders a 64 $\times$ 64 image for 6 important body regions during optimization. However, limited by the computation needs of Mip-NeRF-360, it can only produce low-resolution avatars without high-resolution details. Also, DreamHuman use SMPL shape for direct geometric supervision, which tends to provide skin-tight avatars. Our method, on the other hand, is more controllable and flexible, allowing for the creation of a wider range of accessories, clothing, and other features. Our AvatarVerse introduces a progressive high-resolution generation strategy. This involves gradually decreasing the camera's radius and focusing on distinct body parts, which facilitates the creation of a diverse range of accessories, clothing, and other elements. Our use of progressive grid also ensures a fine-grained generation.

\begin{figure*}[h]
\begin{center}
\includegraphics[width=1.0\linewidth]{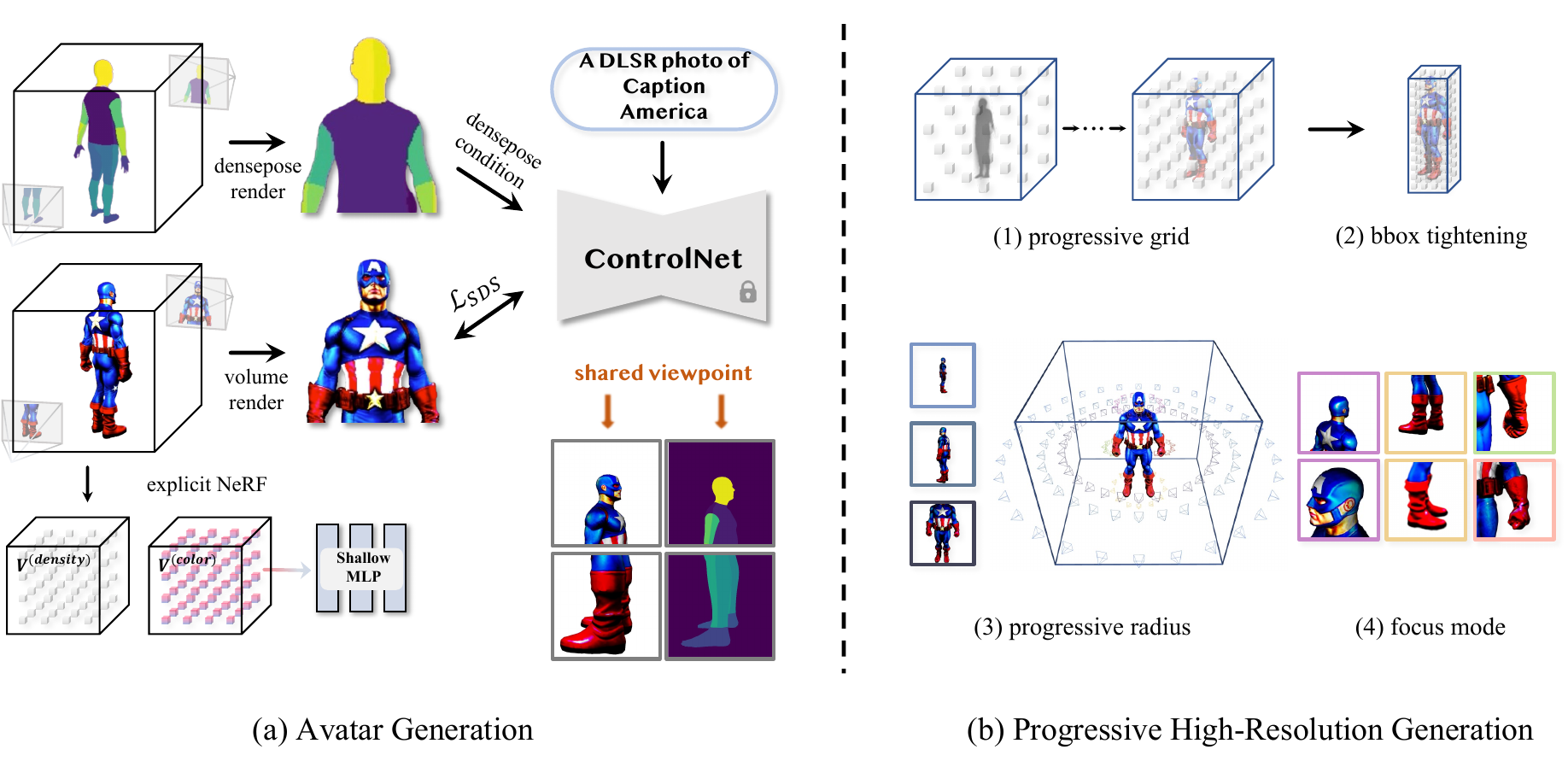}
\end{center}
\caption{The overview of AvatarVerse. Our network takes a text prompt and DensePose signal as input to optimize an explicit NeRF via a DensePose-COCO pre-trained ControlNet. We use strategies including progressive grid, progressive radius, and focus mode to generate high-resolution and high-quality 3D avatars.}
\label{fig:method}
\end{figure*}

\begin{figure*}[h]
\begin{center}
\includegraphics[width=1.0\linewidth]{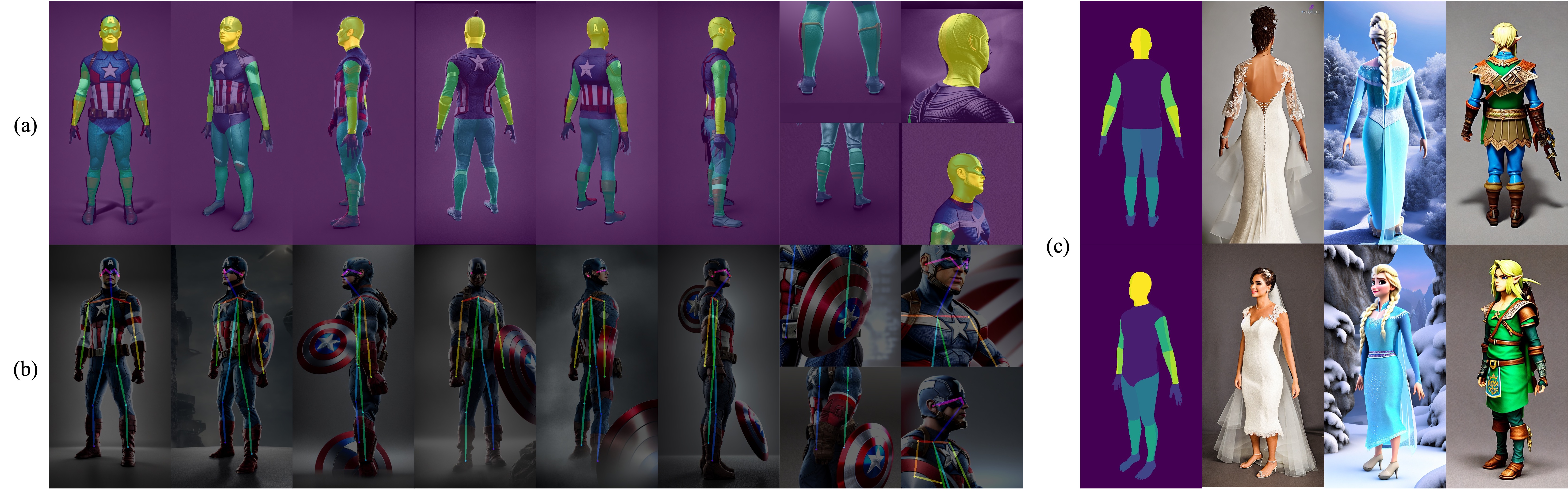}
\end{center}
\caption{Qualitative results of our DensePose-conditioned ControlNet. (a) 10 generated images controlled by DensePose with varying viewpoints and body parts. (b) 10 corresponding images with the same viewpoints controlled by human pose (Openpose) signals. It often fails to generate the backside of the avatar ($4$-th (b)) and struggles with part generation (the last two columns). (c) non-skin-tight generation results in both realistic and fictional avatars.}
\label{fig:control-comparison}
\end{figure*}

\section{3. Methodology} \label{sec: 3}

In this section, we present AvatarVerse, a fully automatic pipeline that can make a realistic 3D avatar from nothing but a text description and a body pose. 
After introducing some preliminaries, we first explain the DensePose-conditioned SDS loss, which
facilitates pose-aware 3D avatar synthesis and effectively mitigates the Janus problem. 
We then introduce novel strategies that enhance the synthesis quality: the progressive high-resolution generation strategy and the avatar surface smoothing strategy.

\subsection{3.1. Preliminaries}

(1) \textit{Score Distillation Sampling}, first proposed by DreamFusion \cite{Poole2022DreamFusionTU}, distills the prior knowledge from a pretrained diffusion model $\boldsymbol{\epsilon}_\phi$ into a differentiable 3D representation $\theta$. Given a rendered image $x = g(\theta)$ from the differentiable NeRF model $g$, we add random noise $\boldsymbol{\epsilon}$ to obtain a noisy image. SDS then calculates the gradients of parameter $\theta$ by minimizing the difference between the predicted noise $\boldsymbol{\epsilon}_\phi\left(x_t ; y, t\right)$ and the added noise $\boldsymbol{\epsilon}$: 
\begin{equation}\label{eq:1}
\nabla_\theta \mathcal{L}_{\text{SDS}}\left(\phi, x_\theta\right)=\mathbb{E}_{t, \boldsymbol{\epsilon}}\left[w(t)\left(\boldsymbol{\epsilon}_\phi\left(z_t ; y, t\right)-\boldsymbol{\epsilon}\right) \frac{\partial x}{\partial \theta}\right],
\end{equation}
where $z_t$ denotes the noisy image at noise level $t$, $w(t)$ is a weighting function that depends on the noise level $t$ and the text prompt $y$.

(2) \textit{SMPL} \cite{Loper2015SMPLAS, Bogo2016KeepIS} is a 3D parametric human body model. It contains 6,890 body vertices and 24 keypoints. By assembling pose parameters $\xi \in \mathbb{R}^{K \times 3}$ and body shape parameter $\beta \in \mathbb{R}^{10}$, the 3D SMPL model can be represented by:
\begin{equation}\label{eq:2}
T(\beta, \xi)=\bar{T}+B_S(\beta)+B_P(\xi)
\end{equation}
\begin{equation}\label{eq:3}
M(\beta, \xi)=\operatorname{LBS}\left(T(\beta, \xi), J(\beta), \xi, \mathcal{W}\right),
\end{equation}
where $T(\beta, \xi)$ denotes the non-rigid deformation combining the mean template shape $\bar{T}$ from the canonical space, the shape-dependent deformations $B_S(\beta) \in \mathbb{R}^{N \times 3}$ and the pose-dependent deformations $B_P(\xi) \in \mathbb{R}^{N \times 3}$. $\operatorname{LBS}(\cdot)$ represents the linear blend skinning function corresponding to articulated deformation. It maps $T(\beta, \xi)$ based on the corresponding keypoint positions $J(\beta) \in \mathbb{R}^{N \times 3}$, pose $\xi$ and blend weights $\mathcal{W} \in \mathbb{R}^{N \times K}$. The body vertex $\mathbf{v}_o$ under the observation pose is 
\begin{equation}\label{eq:4}
\mathbf{v}_o=\sum_{k=1}^K w_k \mathcal{G}_k\left(\xi, j_k\right),
\end{equation}
where $w_k$ is the skinning weight, $\mathcal{G}_k\left(\xi, j_k\right)$ is the affine deformation transforms the $k$-th joint $j_k$ from canonical space to the observation space.

(3) \textit{DensePose} \cite{Gler2018DensePoseDH} is a pioneering technique that facilitates the establishment of dense correspondences between a 2D image and a 3D, surface-based model of the human body.
Leveraging the SMPL model \cite{Loper2015SMPLAS}, DensePose can assign each triangular face within the SMPL mesh to one of the 24 pre-defined body parts. This correspondence allows for the generation of part-labeled 2D body images from any given viewpoint by rendering the associated regions from the SMPL mesh.

\subsection{3.2. DensePose SDS Loss}

Prior research \cite{Poole2022DreamFusionTU, Lin2022Magic3DHT} predominantly employs supplementary text prompts, such as ``front view'' or ``overhead view'', to enhance view consistency. However, reliance solely on text prompts proves inadequate for accurately conditioning a 2D diffusion model on arbitrary views. This inadequacy engenders instability in 3D model synthesis, giving rise to issues like the Janus problem. As a solution, we propose the utilization of DensePose \cite{Gler2018DensePoseDH} as a more robust control signal, as depicted in Figure \ref{fig:method}.

We choose DensePose as the condition because it delivers precise localization of 3D body parts in 2D images, affording intricate details and boundary conditions that may be overlooked by skeletal or other types of conditions. Notably, it exhibits resilience in challenging scenarios, facilitating accurate control even when body parts are partially concealed. 

We first train a ControlNet \cite{Zhang2023AddingCC} conditioned by DensePose part-labeled annotations using the DeepFashion \cite{Liu2016DeepFashionPR} dataset. Figure \ref{fig:control-comparison} illustrates the capabilities of our ControlNet in generating high-quality view-consistent images, including various viewpoints and body parts such as full body, legs, head, and more. 
Given a specific camera viewpoint and pose $P$, we generate the DensePose condition image $c$ by rendering the part-labeled SMPL model with the corresponding pose $P$. The conditioned SDS loss is shown in the following equation:
\begin{equation} \label{eq:5}
\small{
    \nabla_\theta \mathcal{L}_{\text{\textit{P}-SDS}}\left(\phi, x = g(\theta, P)\right)=\mathbb{E}_{t, \boldsymbol{\epsilon}}\left[w(t)\left(\boldsymbol{\hat{\epsilon}}-\boldsymbol{\epsilon}\right) \frac{\partial x}{\partial \theta}\right]
    }
\end{equation}

\begin{equation} \label{eq:6}
\boldsymbol{\hat{\epsilon}} = \boldsymbol{\epsilon}_\phi\left(z_t ; y, t, c = h(\text{SMPL}, P)\right)
\end{equation}

Here, $g$ and $h$ represent the NeRF render function and SMPL render function, respectively. 
The NeRF model and the SMPL pose model share identical camera viewpoints. This alignment of viewpoints enables coherent and consistent representations between the scene captured by NeRF and the corresponding human pose modeled by SMPL, allowing for better avatar generation. Our DensePose-conditioned ControlNet can generate various non-skin-tight realistic and fictional avatars as shown in Figure \ref{fig:control-comparison} (c).

\begin{figure*}[!htbp]
\begin{center}
\includegraphics[width=1.0\linewidth]{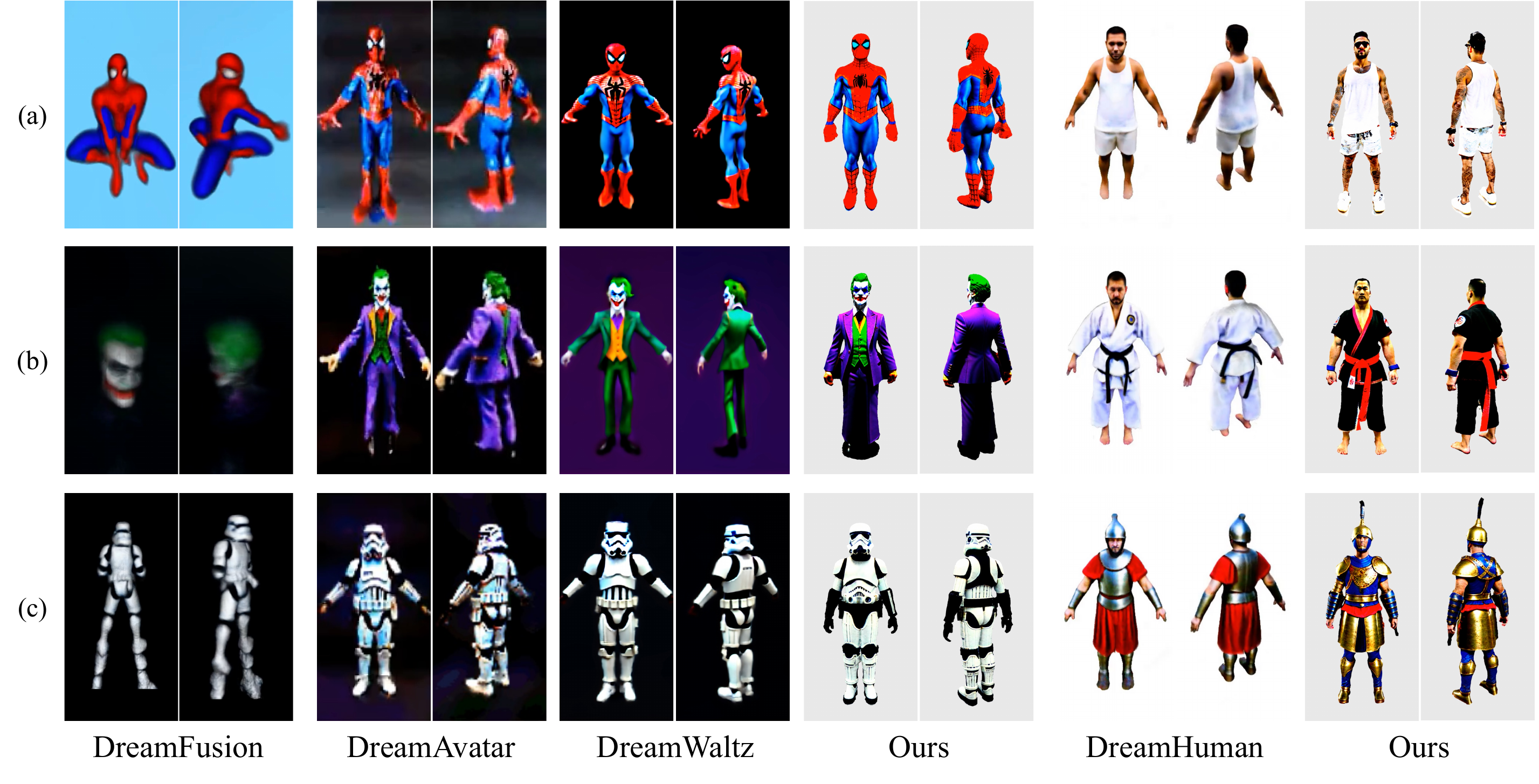}
\end{center}
\caption{Qualitative comparisons with four SOTA methods. We show several non-cherry-picked results generated by AvatarVerse. Our method generates higher-resolution details and maintains a fine-grained geometry compared with other methods. (a): "Spiderman"; "
a man wearing a white tanktop and shorts", (b): "Joker"; "a karate master wearing a Black belt", (c): "Stormtrooper"; "a Roman soldier wearing his armor".}
\vspace{-0.5em}
\label{fig:sota}
\end{figure*}

\subsection{3.3. Progressive High-Resolution Generation}

Previous studies commonly apply SDS loss over the entire body, such global guidance often fails to produce high-quality details, especially for areas like hands, face, etc. These approaches lack effective guidance mechanisms to ensure the generation of high-quality, detailed geometry and textures. To address this limitation, we propose a variety of guidance strategies aimed at promoting the generation of accurate and detailed representations, including progressive grid, focus mode, and progressive radius.

\subsubsection{Progressive grid}
Progressive training strategy is commonly used in 2d generation and 3d reconstruction method \cite{Karras2019AnalyzingAI, Liu2020NeuralSV, Sun2021DirectVG}, while we find
it critical in our method for neat and efficient 3d avatar generation.
We set a predetermined number of voxels $N_v$ as the final model resolution and double the voxel number after certain steps of optimization. The voxel size $s_v$ is updated accordingly. 
During the early stage of training, we only need to generate a rough avatar shape. By allocating fewer grids, we can reduce the learning space and minimize floating artifacts.
This strategy enables a gradual refinement of the avatar throughout the optimization process, allowing the model to adaptively allocate computational resources.

Also, the early stage of NeRF optimization is dominated by free space (i.e., space with low density). Motivated by this fact, we aim to find the areas of coarse avatar and allocate computational and memory resources to these important regions. To delineate the targeted area, we employ a density threshold to filter the scene and use a bounding box (bbox) to tightly enclose this area.

Let $d_x$, $d_y$, $d_z$ represent the lengths of the tightened bbox, he voxel size can be computed as $s_v = \sqrt[3]{\frac{d_x \times d_y \times d_z}{N_v}}$. By shrinking the lengths of the bbox, the voxel size decreases, enabling high-resolution and more voxel around the avatar. This would enhance the model's ability to capture and model intricate details, such as fine-grained body contours, facial features, and clothing folds.

\subsubsection{Progressive Radius}
 Let \texttt{pg\_ckpt} be the set of checkpoint steps. When reaching the training step in \texttt{pg\_ckpt}, we decrease the radius of the camera by $20\%$. This allows for gradual rendering of finer details stage by stage. By applying the conditioned SDS loss to smaller regions of the avatar, the model can capture and emphasize intricate features, ultimately producing more realistic and visually appealing outputs.

\subsubsection{Focus Mode} 
Similarly, to generate better intricacy in specific body parts, we introduce a focus mode (as illustrated in Fig. \ref{fig:method} (b)) during both the coarse stage and fine stage. 
Thanks to the SMPL prior, we can easily compute the raw body parts positions for any given pose. By placing the camera close to important body parts, loss calculation can be performed in a very small avatar region with 512 $\times$ 512 resolution. Owing to the stable performance of our DensePose ControlNet, as shown in Fig. \ref{fig:method}, partial body can be generated without additional computational resources. Focus mode can thus facilitate the creation of high-quality avatar details.

\subsubsection{Mesh Refinement}
To render fine-grained high-resolution avatars within reasonable memory constraints and computation budgets, we further incorporate deformable tetrahedral grids \cite{Lin2022Magic3DHT, Shen2021DeepMT} to learn textured 3D meshes of the generated avatars. Similar to \cite{Lin2022Magic3DHT}, we use the trained explicit NeRF as the initialization for the mesh geometry, and optimize the mesh via backpropagation using the DensePose conditioned SDS gradient (Eq. \ref{eq:5}).

\subsection{3.4. Avatar Surface Smoothing} 

Maintaining a globally coherent avatar shape for explicit grids during optimization can be challenging due to the high degree of freedom and lack of spatial coherence. Individual optimization of each voxel point limits information sharing across the grid, resulting in a less smooth surface for the generated avatar and some local minima.

To address this problem, we follow the definition of the Gaussian convolution $\mathcal{G}$ in \cite{Wu2022VoxurfVE} and include a modified smoothness regularization formulated as:

\begin{equation}
    \mathcal{L}_{\text{smooth}}(V)=\left\|\mathcal{G}\left(V, k_g, \sigma_g\right)-V\right\|_2^2
\end{equation}

Here, $k_g$ represents the kernel size, and $\sigma_g$ represents the standard deviation. We apply this smoothness term to the gradient of the density voxel grid, resulting in a gradient smoothness loss $\mathcal{L}_{\text{smooth}}(\nabla V^{(\text{density})})$. This encourages a smoother surface and mitigates the presence of noisy points in the free space. The overall loss of our approach is defined as follows, with $\lambda$ representing the smoothness coefficient:

\begin{equation}
    \mathcal{L}= \mathcal{L}_{\text{\textit{P}-SDS}} + \lambda * \mathcal{L}_{\text{smooth}}(V)
\end{equation}

\section{4. Experiments} \label{sec: 4}

In this section, we illustrate the effectiveness of our proposed method. We demonstrate the efficacy of each proposed strategy and provide a detailed comparison against recent state-of-the-art methods.

\subsection{4.1. Implementation Details} 
We follow \cite{Sun2021DirectVG} to implement the explicit NeRF in our method. For each text prompt, we train AvatarVerse for 5000 and 4000 iterations in the coarse stage and mesh refinement stage, respectively. The whole generation process takes around 2 hours on one single NVIDIA A100 GPU. We include initialization, densepose training and progressive high-resolution generation details in this section. For more comprehensive experiment details, we refer the reader to our Supplementary Material.

\subsubsection{Initialization}
To aid in the early stages of optimization, we adopt a technique inspired by \cite{Poole2022DreamFusionTU} and introduce a small ellipsoidal density "blob" around the origin. The dimensions of the "blob" in the XYZ axes are determined based on the range of coordinates in the SMPL pose model. Furthermore, we incorporate additional SMPL-derived density bias \cite{Cao2023DreamAvatarTG} to facilitate avatar generation.

\subsubsection{DensePose Training}
We annotate the DeepFashion dataset \cite{Liu2016DeepFashionPR} using a pretrained DensePose \cite{Gler2018DensePoseDH} model, resulting in over 800K image pairs. The ControlNet is trained using these image pairs with BLIP2-generated text prompt \cite{li2023blip2}. The diffusion model employed in our approach is SD1.5.

\subsubsection{Progressive High-Resolution Generation}

For the progressive grid, we double the number of voxels at 500, 1500, and 2000 iterations at the coarse stage. After 3000 steps in the coarse stage, we shrink the bounding box to the region where the density exceeds 0.1. Our progressive radius consists of three stages, where the camera radius ranges from 1.4 to 2.1, 1 to 1.5, and 0.8 to 1.2 respectively. We reduce the radius at 1000 and 2000 iterations across both stages. Our focus mode starts from the 1000\texttt{th} step in the coarse stage and is consistently employed throughout the mesh refinement phase.

\subsection{4.2. Qualitative Results}

\subsubsection{Comparison with SOTA methods}

We present qualitative comparisons with DreamFusion \cite{Poole2022DreamFusionTU}, DreamAvatar \cite{Cao2023DreamAvatarTG}, DreamWaltz \cite{Huang2023DreamWaltzMA}, and DreamHuman \cite{Kolotouros2023DreamHumanA3} in Fig. \ref{fig:sota}. Our method consistently outperforms these approaches in terms of both geometry and texture quality. The surface of the avatars generated by our method is exceptionally clear, owing to our progressive high-resolution generation strategy. In comparison to DreamHuman, the avatars produced by our method exhibit a richer array of details across all cases, encompassing skin, facial features, clothing, and more.

\begin{figure}[!htbp]
\begin{center}
\includegraphics[width=1.0\linewidth]{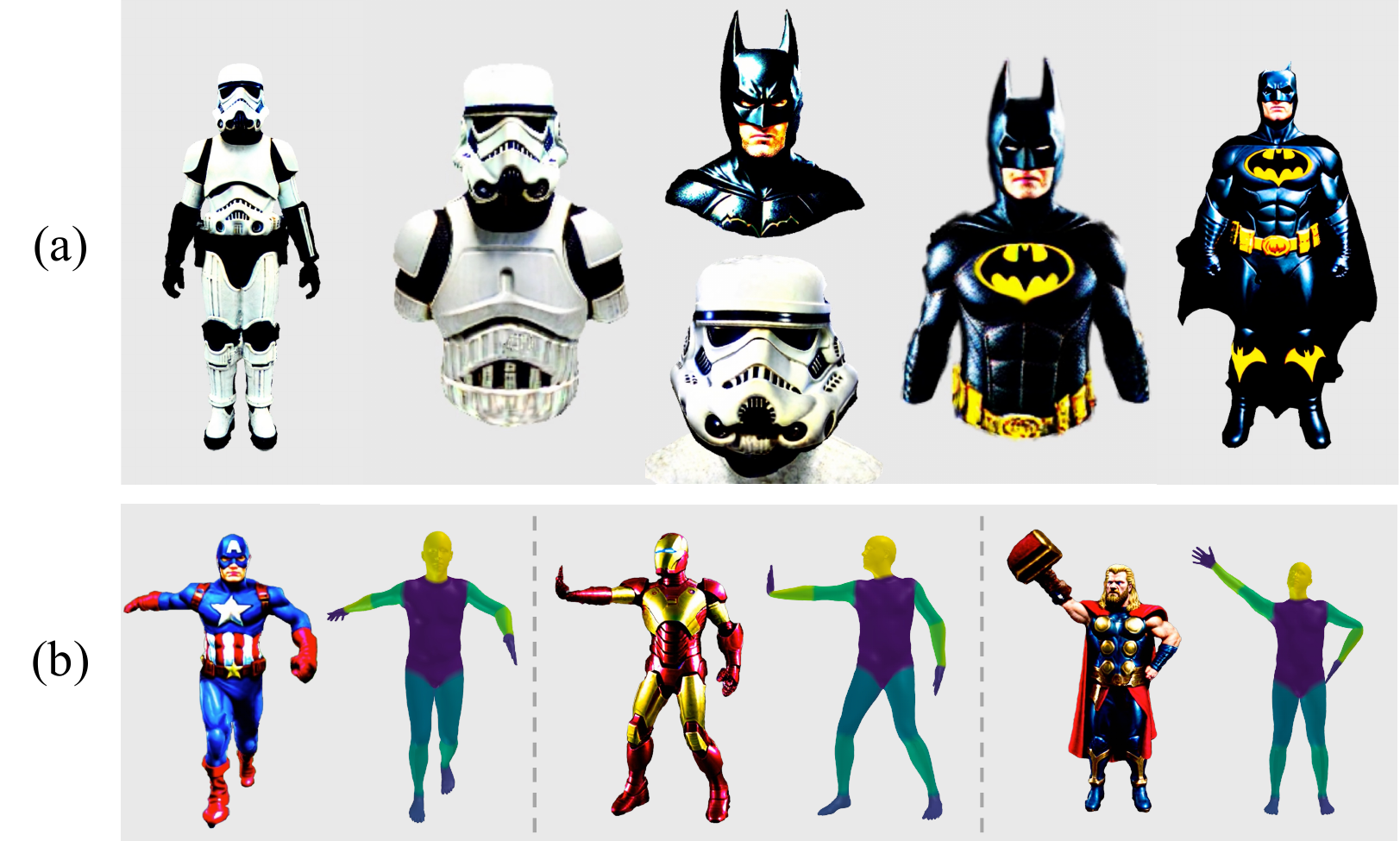}
\end{center}
\caption{Flexible Avatar Generation. (a) Partial Generation. All results are generated with the same text prompt "Stormtrooper" and "Batman". (b) Arbitrary Pose Generation.}
\vspace{-1em}
\label{fig:partial}
\end{figure}

\subsubsection{Flexible Avatar Generation}
In Fig. \ref{fig:partial}, we demonstrate the capability of our method in generating 3D partial avatars, which is not achievable by other existing methods due to the absence of the DensePose control. 
Our method enables the partial generation by directly modifying the input DensePose signal, eliminating the need for additional descriptive information such as "The head of..." or "The upper body of...". This allows us to generate partial avatars of various types thanks to the attached semantics, including full-body, half-body, head-only, hand-only, and more. Additionally, our AvatarVerse is capable of generating avatars in various poses, showcasing our stable control over view consistency.

\subsection{4.3. User Study}

\begin{figure}[!htbp]
\begin{center}
\includegraphics[width=1.0\linewidth]{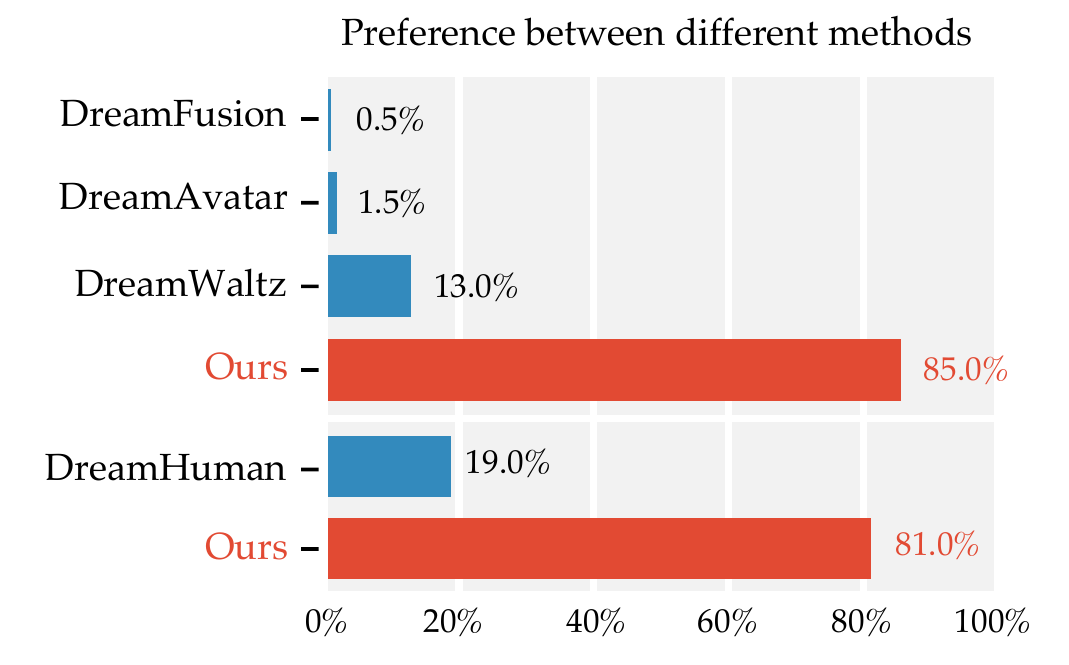}
\end{center}
\vspace{-0.5cm}
\caption{Quantitative results of user study.}
\label{fig:userstudy}
\end{figure}

To further assess the quality of our generated 3D avatars, we conduct user studies comparing the performance of our results with four SOTA methods under the same text prompts. We randomly select 30 generated outcomes (presented as rendered rotating videos) and ask 16 volunteers to vote for their favorite results based on geometry and texture quality.
In Fig. \ref{fig:userstudy}, we compare AvatarVerse with DreamFusion \cite{Poole2022DreamFusionTU}, DreamAvatar \cite{Cao2023DreamAvatarTG}, and DreamWaltz \cite{Huang2023DreamWaltzMA}, demonstrating a significant preference for our method over the other three approaches.

We also compare our method with DreamHuman \cite{Kolotouros2023DreamHumanA3} in terms of realistic human. A remarkable $81\%$ of volunteers voted in favor of our AvatarVerse.

\subsection{4.4. Ablation Study}

\subsubsection{Effectiveness of Progressive Strategies}

\begin{figure}[h]
\begin{center}
\includegraphics[width=1.0\linewidth]{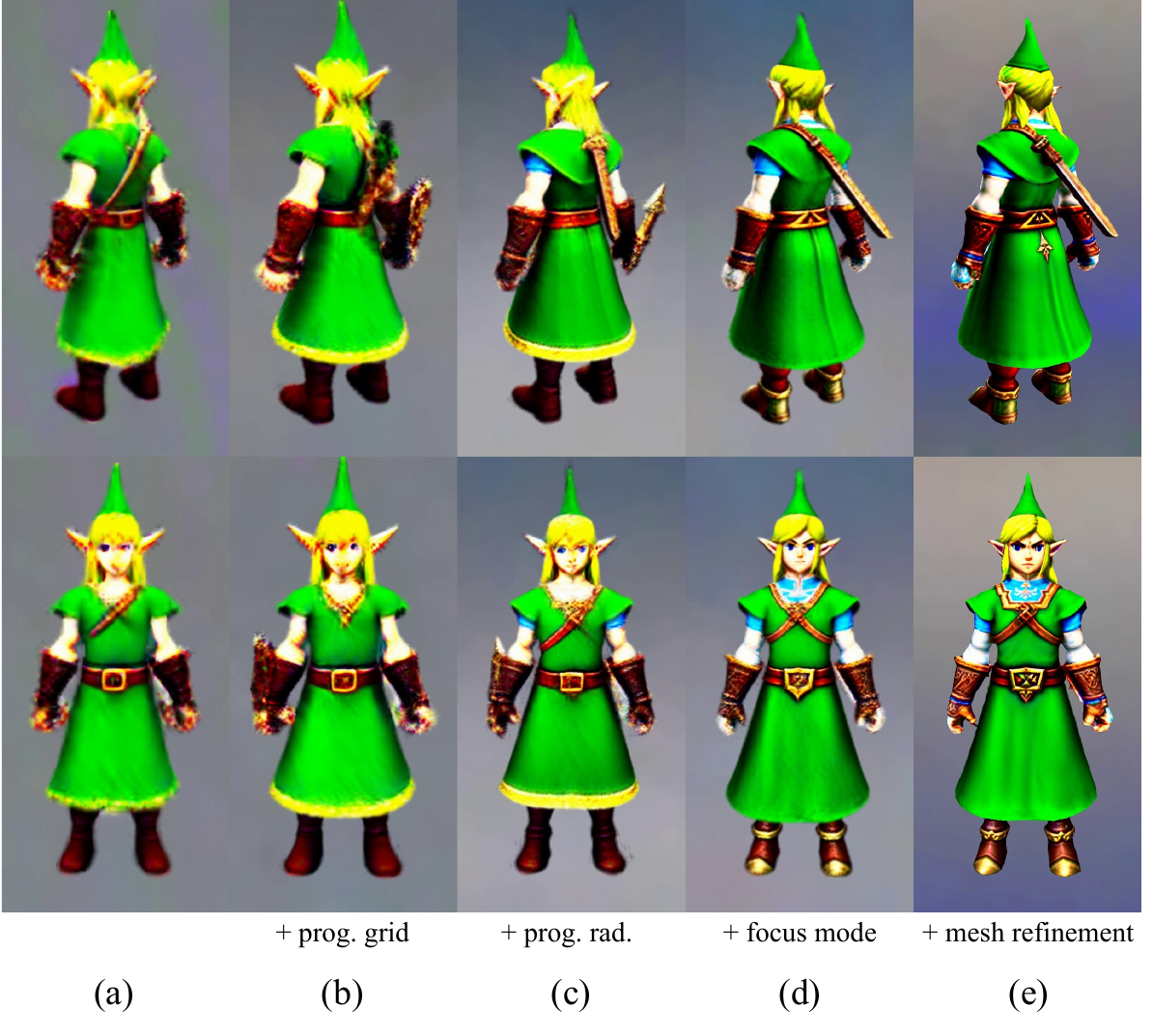}
\end{center}
\caption{Impact of progressive strategies. (a) none progressive strategy; (b) add progressive grid; (c) add progressive radius upon (b); (d) add focus mode upon (c); (e) add mesh refinement, our full method.}
\vspace{-1em}
\label{fig:ablation}
\end{figure}

To evaluate the design choices of AvatarVerse, we conduct an ablation study on the effectiveness of b) the progressive grid, c) the progressive radius, d) the focus mode, and e) the mesh refinement. We sequentially add these components and report the results in Fig. \ref{fig:ablation}. The initial result lacks detail (e.g., no sword in the back, no armguards) and exhibits numerous floating artifacts. The overall quality is blurry and unclear.
Upon incorporating the progressive grid, more voxels are gathered around the avatar region, this introduces more details into the avatar. 
By progressively narrowing the camera distance, the model can leverage the detail inherent in the latent diffusion, thereby eliminating a large number of floating artifacts and enhancing local details, such as the sword in the back.
The focus mode further zooms in and utilizes a resolution of 512 $\times$ 512 to target and optimize certain body parts, generating high-definition and intricate local details. The mesh refinement further optimize 3D mesh of the coarse avatar, resulting in finer avatar texture.

\begin{figure}[h]
\begin{center}
\includegraphics[width=1.0\linewidth]{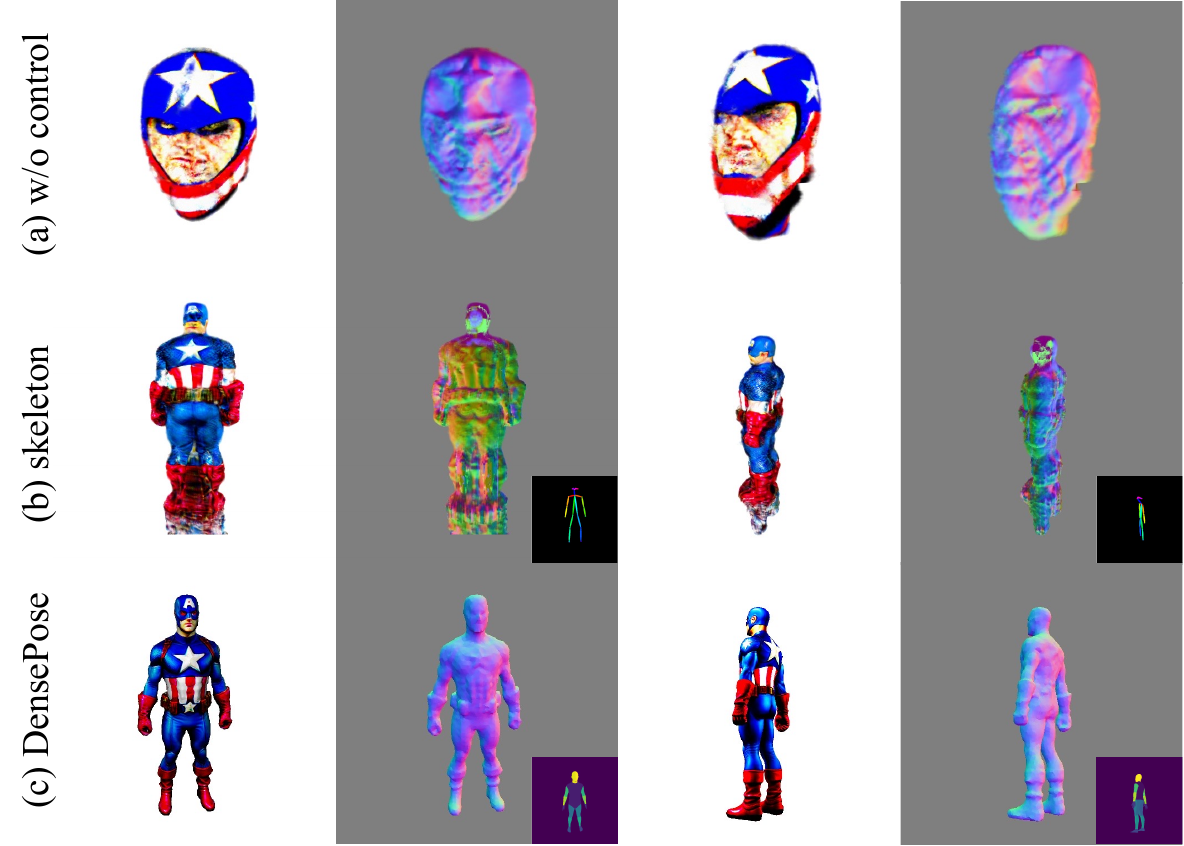}
\end{center}
\caption{Impact of control signal. (a) without additional control; (b) with skeleton control; (c) with our DensePose control. For each type, we show the RGB, normal, depth, and the corresponding control signal.}
\vspace{-1em}
\label{fig:control}
\end{figure}

\subsubsection{Effectiveness of DensePose Control} 
Figure \ref{fig:control} illustrates the influence of various control signals. When conditioned by the skeleton, the model can generate avatars that more closely resemble human figures. However, the avatar's edges appear blurry and still face severe Janus problem. By incorporating DensePose control into our framework, we achieve more precise avatar boundaries, intricate details, and stable avatar control, resulting in a substantial improvement in the overall quality and appearance of the generated avatars.

\subsubsection{Effectiveness of Surface Smoothing}
\begin{figure}[H]
\begin{center}
\includegraphics[width=1.0\linewidth]{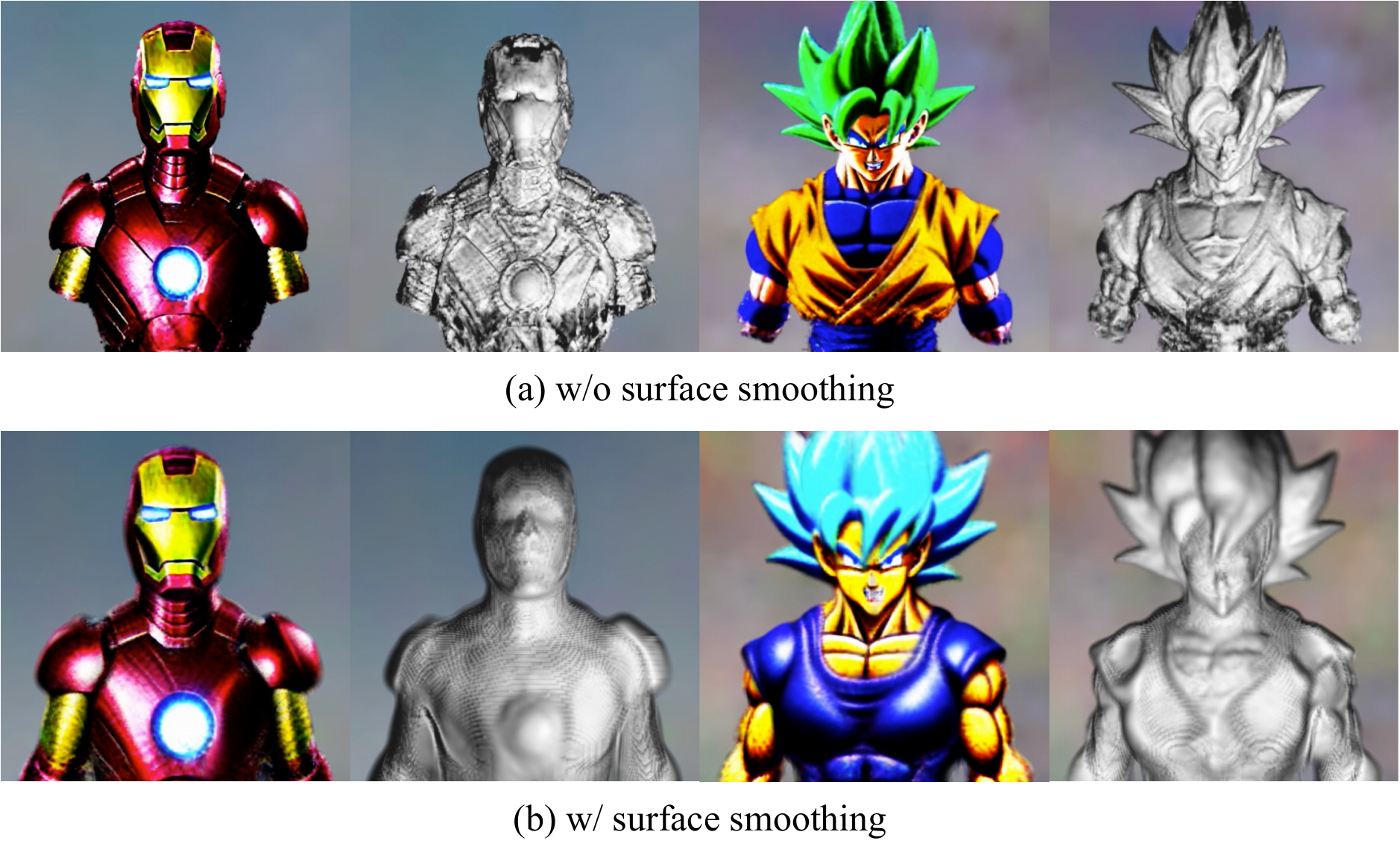}
\end{center}
\caption{Impact of surface smoothing strategy. (a) without surface smoothing; (b) with surface smoothing. Results are generated with the same text prompt.}
\vspace{-1em}
\label{fig:smooth}
\end{figure}

 Avatar surface smoothing plays a critical role in the AvatarVerse framework, as it guarantees the generated avatars exhibit compact geometry and smooth surfaces. As shown in Figure \ref{fig:smooth}, by finding a balance between the smooth loss and the conditioned SDS loss, the visual quality and realism of the avatars are greatly improved.

\section{Conclusion}

In this paper, we introduce AvatarVerse, a novel framework designed to generate high-quality and stable 3D avatars from textual prompts and poses. 
By employing our trained DensePose-conditioned ControlNet, we facilitate stable partial or full-body control during explicit NeRF optimization. Our 3D avatar outcomes exhibit superior texture and geometry quality, thanks to our progressive high-resolution generation strategy. Furthermore, the generated avatars are easily animatable through skeletal binding, as they exhibit high alignment with the joints of the SMPL model. Through comprehensive experiments and user studies, we demonstrate that our AvatarVerse significantly outperforms previous and contemporary approaches. We believe that our approach renews the generation of high-quality 3D avatars in the neural and prompt-interaction era.

\newpage
\bibliography{ref.bib}

\end{document}